\documentclass[sigconf]{acmart}  
\usepackage{hyperref}
\hypersetup{
    colorlinks=true,
    linkcolor=blue
}
\usepackage{bm}
\usepackage{enumitem}
\DeclareMathAlphabet{\altmathcal}{OMS}{cmsy}{m}{n}
\usepackage[ruled, noend]{algorithm2e}
\usepackage{graphicx}
\usepackage{textcomp}
\usepackage{xcolor}
\usepackage{lipsum}
\usepackage{makecell}

\usepackage{multirow}

\newcommand{\dataset}{\textsc{GUIDE}}
\newcommand{\methodshort}{\textsc{CGR}}
\newcommand{\method}{\textsc{Copilot Guided Response}}
\newcommand{\methodlong}{\textsc{Microsoft Copilot for Security Guided Response}}

\newcommand{\hide}[1]{}

\AtBeginDocument{%
  }

\setcopyright{acmlicensed}
\copyrightyear{2024}
\acmYear{2024}
\acmDOI{XXXXXXX.XXXXXXX}

\acmISBN{978-1-4503-XXXX-X/18/06}

\begin{document}

\title{AI-Driven Guided Response for Security Operation Centers with Microsoft Copilot for Security}

\author{Scott Freitas}
\email{scottfreitas@microsoft.com}
\affiliation{%
  \institution{Microsoft Security Research}
  \city{Redmond}
  \state{WA}
  \country{USA}
}

\author{Jovan Kalajdjieski}
\email{jovank@microsoft.com}
\affiliation{%
  \institution{Microsoft Security Research}
  \city{Redmond}
  \state{WA}
  \country{USA}
}

\author{Amir Gharib}
\email{agharib@microsoft.com}
\affiliation{%
  \institution{Microsoft Security Research}
  \city{Redmond}
  \state{WA}
  \country{USA}
}

\author{Robert McCann}
\email{robert.mccann@microsoft.com}
\affiliation{%
  \institution{Microsoft Security Research}
  \city{Redmond}
  \state{WA}
  \country{USA}
}

\renewcommand{\shortauthors}{Freitas \& Kalajdjieski et al.}

\begin{abstract}
Security operation centers contend with a constant stream of security incidents, ranging from straightforward to highly complex.
To address this, we developed \methodlong{} (\methodshort{}), an industry-scale ML architecture that guides security analysts across three key tasks---%
(1) \textit{investigation}, providing essential historical context by identifying similar incidents;
(2) \textit{triaging} to ascertain the nature of the incident---whether it is a true positive, false positive, or benign positive;
and (3) \textit{remediation}, recommending tailored containment actions.
\methodshort{} is integrated into the Microsoft Defender XDR product and deployed worldwide, 
generating millions of recommendations across thousands of customers. 
Our extensive evaluation, incorporating internal evaluation, collaboration with security experts, and customer feedback, demonstrates that \methodshort{} delivers high-quality recommendations across all three tasks.
We provide a comprehensive overview of the \methodshort{} architecture, setting a precedent as the first cybersecurity company to openly discuss these capabilities in such depth.
Additionally, we release \href{https://www.kaggle.com/datasets/Microsoft/microsoft-security-incident-prediction}{\dataset{}}, the largest public collection of real-world security 
incidents, spanning 13M evidences across 1M incidents annotated with ground-truth triage labels by customer security analysts.
This dataset represents the first large-scale cybersecurity resource of its kind, supporting the development and evaluation of guided response systems and beyond.

\end{abstract}




\maketitle

\section{Introduction}\label{sec:introduction}
In the rapidly evolving cybersecurity landscape, the sharp rise in threat actors has overwhelmed enterprise security operation centers (SOCs) with an unprecedented volume of incidents to triage~\cite{forrester2020state}. 
This surge requires solutions that can either partially or fully automate the remediation process. 
Fully automated systems demand an exceptionally high confidence threshold (e.g., 99\%) to ensure correct actions are taken to avoid inadvertently disabling critical enterprise assets. 
Consequently, attaining such a high level of confidence often renders full automation impractical.

\begin{figure*}[t!]
    \centering
    \includegraphics[width=\textwidth]{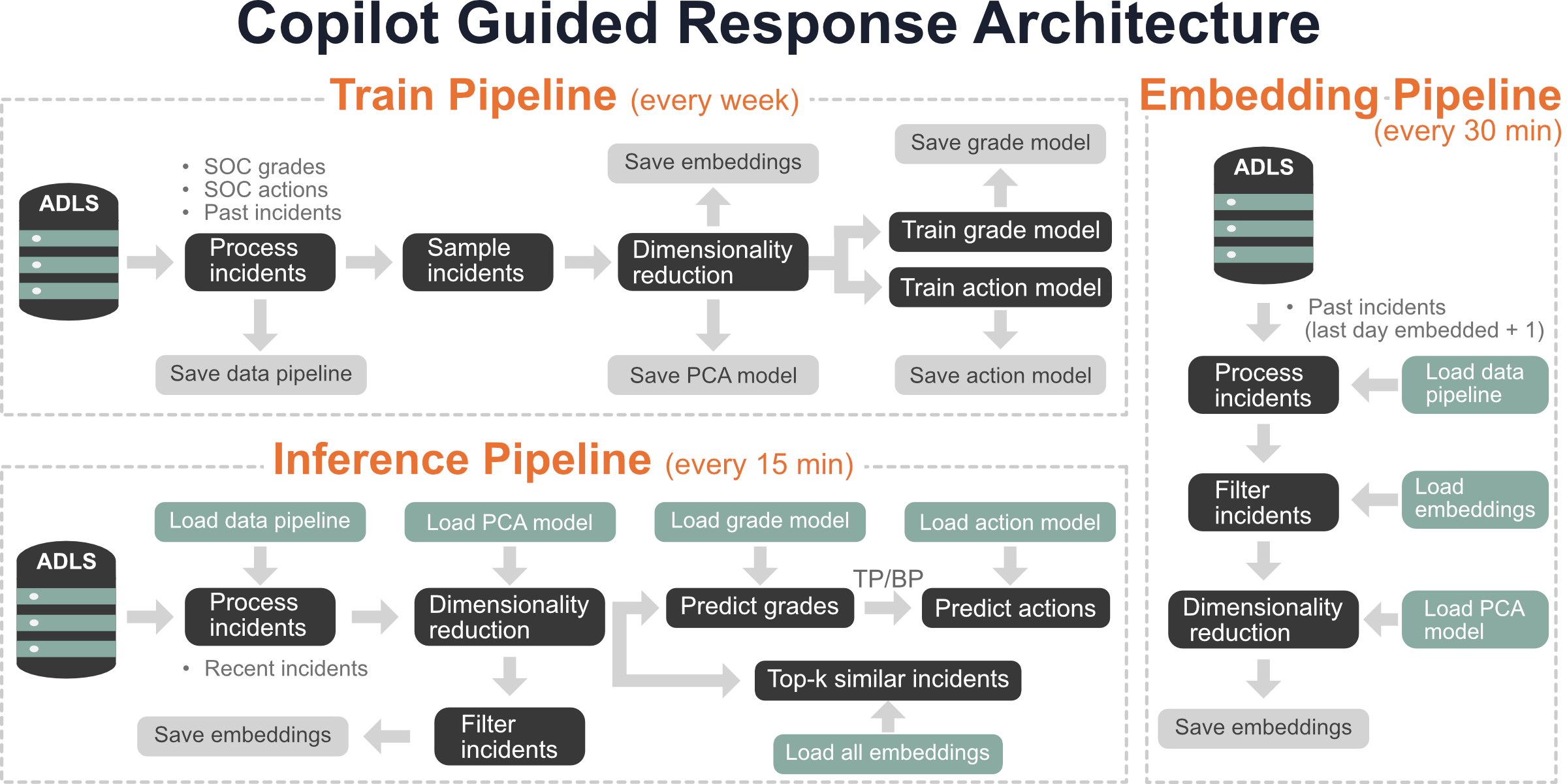}
    \caption{Overview of the \method{} architecture.
    Train Pipeline: Running weekly, this process trains grade and action recommendation models based on historical SOC telemetry.
    Inference Pipeline: Running every 15 minutes, this process generates grade, action, and similar incident recommendations for incoming incidents by leveraging the models created in the train pipeline.
    Embedding Pipeline: Running every 30 minutes until 180 days of historical embeddings exist, this job creates historical embeddings of SOC incidents for the similar incident recommendation algorithm in the inference pipeline.
    }
    \label{fig:crown}
\end{figure*}

This challenge has catalyzed the development of guided response (GR) systems to support SOC analysts by facilitating informed decision-making.
Extended Detection and Response (XDR) products are ideally positioned to deliver precise, context-rich guided response recommendations thanks to their comprehensive visibility across the entire enterprise security landscape.
By consolidating telemetry across endpoints, network devices, cloud environments, email systems, and more, XDR systems can harness a wide array of data to provide historical context, generate detailed insights into the nature of threats, and recommend tailored remediation actions.

\vspace{2mm}\noindent
\textbf{Guided response challenges.}
Scalable and accurate GR systems face several key challenges that require a combination of innovative ML system design, and a deep understanding of cybersecurity:

\begin{enumerate}[topsep=4pt, leftmargin=*, itemsep=3pt]

    \item \textbf{Complexity of security incidents.} The extensive variety of security products, each with thousands of custom and built-in detectors, creates a complex incident landscape further compounded by a scarcity of labeled data.

    \item \textbf{High precision and recall.} 
    Analysts require reliable guidance, necessitating systems that deliver high precision and recall across investigation, triaging, and remediation tasks.

    \item \textbf{Scalable architecture.} 
    Generating recommendations at the million-scale across terabytes of data requires a robust and scalable ML architecture.

    \item \textbf{Adaptive to unique SOC preferences.} The system must be able to adapt to specific operational workflows, product configurations, and detection logic of individual SOCs.

    \item \textbf{Continuous learning and improvement.} To remain effective against evolving cyber threats and changes in the security product landscape, the system must continuously learn and improve autonomously.
\end{enumerate}

\subsection{Contributions}
We introduce \method{} (Fig.~\ref{fig:crown}), an ML framework designed to tackle guided response at scale.
Our framework makes significant contributions in the following areas:

\begin{itemize}[topsep=2mm, itemsep=0mm, parsep=1mm, leftmargin=*]
    
    \item \textbf{\method{} (\methodshort{}).} 
    The \method{} architecture transforms cybersecurity guided response by detailing the first geo-distributed industry-scale framework capable of processing millions of incidents each day with batch latency of just a few minutes.
    Our ML system scalably delivers three core SOC capabilities---(1) \textit{investigation},
    (2) \textit{triaging}, 
    and (3) \textit{remediation}---%
    seamlessly adapting to a range of scenarios, from single alerts to complex incidents involving hundreds of alerts, where each alert is categorized into one of hundreds of thousands of distinct classes, with new classes continuously added.

    \item \textbf{Largest Cybersecurity Incident Dataset.}
    We introduce \href{https://www.kaggle.com/datasets/Microsoft/microsoft-security-incident-prediction}{\dataset{}}, the largest publicly available collection of real-world cybersecurity incidents under the permissive CDLA-2.0 license. 
    This extensive dataset includes over 13 million pieces of evidence across 1.6 million alerts and 1 million incidents annotated with ground-truth triage labels by customer security analysts, making it an unparalleled resource for the development and evaluation of GR systems and beyond.
    By enabling researchers to study real-world data, \dataset{} advances the state of cybersecurity and supports the development of next-generation ML systems.

    \item \textbf{Extensive Evaluation of \methodshort{}.}
    We provide a comprehensive evaluation of \methodshort{}'s performance, focusing on its industrial applicability rather than comparisons with alternative designs or competing security products, which are often undisclosed. 
    The release of the \dataset{} dataset enables researchers to explore new architectures for maximum performance.
    Our evaluation spans internal testing, expert collaborations, and customer feedback.
    Internal assessments on hundreds of thousands of unseen incidents show triage models achieve 87\% precision and 41\% recall, while action models reach 99\% precision and 62\% recall. 
    Collaboration with Microsoft security experts confirms the efficacy of our similar incident recommendations, with 94\% of incidents deemed relevant, and 98\% of incidents containing one or more recommendations. 
    Customer feedback further underscores its effectiveness, with 89\% of interactions rated positively.

    \item \textbf{Impact to Microsoft Customers and Beyond.} 
    \methodshort{} is integrated into the Microsoft Defender XDR product, a leader in the market~\cite{mellen2024forrester}, and is deployed to hundreds of thousands of organizations worldwide.
    The introduction of \methodshort{} to Microsoft Defender XDR has significantly enhanced the operational capabilities of SOCs by streamlining the decision-making process and providing actionable insights across investigation, triaging, and remediation tasks.
    As a result, Microsoft Defender XDR customers benefit from a more resilient security response, fortified by adaptive ML-driven guided responses that are tailored to the nuances of their specific security environments.
\end{itemize}

\begin{table}[t]
\centering
\begin{tabular}{p{0.08\textwidth} p{0.355\textwidth}}

\textbf{Term} & \textbf{Definition} \\
\cmidrule(r){1-1} \cmidrule(lr){2-2}
Alert & Potential security threat that was detected  \\ \addlinespace
Detector & A security rule or ML model that generates alerts \\ \addlinespace
Entity & File, IP, etc. evidence associated with an alert \\ \addlinespace
Correlation & A link between two alerts based on a shared entity \\ \addlinespace
Incident & Related alerts that are correlated together \\ \addlinespace
SOC & Security operation centers (SOC) protect enterprise organizations from threat actors  \\ \addlinespace
XDR & Extended Detection and Response (XDR) platforms are used by SOCs to protect organizations across the entire enterprise landscape 
\\ \addlinespace
\bottomrule
\end{tabular}
\caption{Terminology and definitions.}
\vspace{-5mm}
\label{table:terminology}
\end{table}

\section{Background}\label{sec:related}
We provide an overview of \textsc{Microsoft Copilot for Security} and review literature relevant to guided response. 
To enhance readability, Table~\ref{table:terminology} details the terminology used in this paper.

\subsection{Microsoft Copilot for Security}
Microsoft Copilot for Security is an AI-driven solution that enhances security professionals' workflows by offering real-time insights and recommendations across Microsoft Defender XDR, Microsoft Sentinel, and Microsoft Intune. 
At launch, it introduced five skills: incident summarization, script analyzer, incident report, Kusto query assistant, and guided response~\cite{what2024caparas}. 
While the first four leverage LLMs with security-specific plugins and post-processing~\cite{overview2024copilot}, guided response includes three machine learning sub-skills---grade recommendation, action recommendation, and similar incident recommendation---tailored to SOC preferences for precise, context-specific insights.

\subsection{Guided Response}
With the market introduction of \methodlong{}, the concept of guided response was formally defined as ``\textit{machine learning capabilities to contextualize an incident and learn from previous investigations to generate appropriate response actions}''~\cite{gali2024triage}. 
Our analysis of academic and industry literature contextualizes relevant contributions within the domain of guided response into three distinct categories: 
(1) investigation that suggests next steps for further analysis;
(2) triaging to determine whether an incident is a true positive, false positive, or  benign positive (e.g., informational);
and (3) remediation which proposes specific response actions to contain and resolve incidents.

\smallskip\noindent
\textbf{Investigation.}
Assisting security analysts in their investigation of incidents is a pivotal aspect of cybersecurity. 
Machine learning assisted investigation typically encompasses: 
(1) similar incident identification~\cite{zhong2018cyber,jiang2024xpert,lin2018data},
(2) investigation assistance~\cite{franco2020secbot, nguyen2021human,perera2019intelligent}, 
and (3) playbook recommendation~\cite{kremer2023ic,kraeva2021application,applebaum2018playbook}. 
Our research focuses on similar incident recommendation, an understudied topic in the context of SOCs. 
While a few industry solutions offer similar incident identification capabilities through methods such as comparing and counting identical artifacts \cite{cortexxsoarincident, cortexxdrincident}, details regarding their methodologies and performance metrics are scarce.
We address this gap by detailing the first industry-scale architecture for similar incident recommendation in SOCs.

\smallskip\noindent
\textbf{Triaging.}
Incident triaging is a vital and time intensive task typically performed by junior analysts to identify incidents requiring further investigation. 
This involves prioritizing incidents for deeper review~\cite{oprea2018made, aminanto2020threat, liu2022rapid, alsubhi2012fuzmet, alsubhi2008alert} and filtering them based on the likelihood of being true versus false positives~\cite{ban2021combat, sopan2018building, hassan2019nodoze, hassan2020tactical}. 
While some industry solutions use ML to automate triaging~\cite{ibmatds2020}, there is limited public information on their architecture or performance. 
Although some companies provide insights~\cite{sopan2018building, feng2017user, oliver2024carbon}, these are often limited to controlled scenarios and lack critical deployment details. 
Our work addresses these gaps by detailing a scalable, adaptable ML triaging architecture that is deployed globally to thousands of SOCs.

\smallskip\noindent
\textbf{Remediation.}
The majority of remediation research centers on intrusion response systems (IRS), designed to notify SOC analysts or dynamically respond to detected intrusions. 
These systems employ various decision-making models, including rule-based approaches~\cite{foo2005adepts}, multi-objective optimization~\cite{li2018dynamic,shameli2016dynamic,toth2002evaluating}, game theory~\cite{huang2015cyber,qin2018risk,zonouz2013rre}, human-in-the-loop~\cite{cortexremediationsuggestion}, reinforcement learning~\cite{alturkistani2022optimizing,stefanova2018off,khoury2020hybrid}, and many others~\cite{bashendy2023intrusion,inayat2016intrusion,shameli2012intrusion} to determine optimal responses based on the system state, nature of the attack, and countermeasure impact.
Despite advancements, there is limited discussion on how to scale these systems to handle industry demands, such as managing millions of incidents, handling complex scenarios with numerous alerts, and customizing responses for SOC preferences. 
Our research bridges this gap, enhancing the scalability and transparency of industry-scale guided remediation systems.

\section{Architecture Overview}\label{sec:architecture}
We detail the \method{} (\methodshort{}) architecture, organized around three key pipelines: train, inference, and embedding, as illustrated in Figure~\ref{fig:crown}. 
We utilize PySpark's distributed computational engine whenever possible, while reserving Python for last mile recommendation tasks that do not have PySpark support. 
Below, we detail how these pipelines synergize to guide security analysts through the processes of investigating, triaging, and remediating security incidents across the enterprise landscape.

\begin{itemize}[topsep=4pt, leftmargin=*, itemsep=3pt]
    \item \textbf{Train pipeline (Section~\ref{sec:train}).} Running weekly, this process trains the grade and action recommendation models using historical SOC telemetry to provide tailored responses.
    This procedure is detailed across ten steps, T1 through T10.

    \item \textbf{Inference pipeline (Section~\ref{sec:inference}).} Operating every 15 minutes, this pipeline generates grade and action recommendations for incoming incidents by leveraging the models developed in the train pipeline. 
    Additionally, it provides similar incident recommendations by matching new incidents with historically similar incidents generated in the embedding pipeline.

    \item \textbf{Embedding pipeline (Section~\ref{sec:embed}).} This pipeline runs every 30 minutes until 180 days of incident embeddings are generated.
    These embeddings form the foundational data that allows the similar incident recommendation algorithm in the inference pipeline to effectively identify similar incidents.
\end{itemize}

\noindent
To adhere to privacy regulations, \methodshort{} is replicated across geographic regions, utilizing Synapse to ensure consistency and compliance. 
Consequently, the following sections focuses on development from the perspective of a single geographic region.

\section{Train Pipeline}\label{sec:train}
\method's training architecture is detailed across two subsections---Section~\ref{subsec:feature} which presents the key steps to collecting and preparing the data; 
and Section~\ref{subsec:train} which discusses the model training and validation process.
A step-by-step overview of the entire training process is provided in Algorithm~\ref{alg:train}.

\subsection{Preprocessing}\label{subsec:feature}
We detail the ten step process (T1-T10) for creating alert and incident dataframes leveraged across all three pipelines.

\medskip\noindent
\textbf{T1---Feature engineering.}
We collect alert telemetry from multiple Azure Data Lake Storage (ADLS) tables and join them into a PySpark alert dataframe.
Each row in the alert dataframe contains columns for unique alert and incident identifiers, complemented by customer-provided grade and remediation action, when available. 
Additionally, each row contains 5 categorical feature columns---OrganizationId, DetectorId, ProductId, Category, and Severity---along with 67 engineered numerical feature columns, developed in close collaboration with Microsoft security research experts. 
We retain rows that lack a customer grade or action, as these alerts can merge with other alerts to form incidents that do contain labels.

\medskip\noindent
\textbf{T2---Feature space compression.}
Before converting the categorical columns to one-hot-encoded representations, we must address the challenge of high cardinality in the DetectorId and OrgId columns. 
In various geographic regions, DetectorIds can exceed 100k and OrgIds can reach up to 50k, creating an extremely large and sparse feature space that often leads to failures during dimensionality reduction in the PySpark cluster. 
To mitigate this, we aggregate the feature space by substituting infrequent values---those associated with fewer than 10 alerts---with a generic value. 
This method, while resulting in some information loss, ensures the system remains within the computational boundaries of the cluster.

\medskip\noindent
\textbf{T3---One-hot-encoding.}
With the preliminary adjustments to our feature space, we can convert all 6 categorical feature columns into their one-hot-encoded (OHE) form. 
This transformation includes key columns such as OrgId, ProductId, and DetectorId, which allows the models to capture SOC-specific tendencies as well as product and detector specific characteristics that evolve over time. 
We bifurcate the data and establish a secondary PySpark alert dataframe that only contains alerts with remediation actions, while retaining all alerts in the original dataframe.
Finally, we store the PySpark OHE pipeline in an Azure Blog Storage container so that it can be used in the inference process to transform the categorical columns.

\medskip\noindent
\textbf{T4---Forming incidents.}
To enhance our ability to make precise incident-level triaging decisions and investigation recommendations, we create a separate incident dataframe.
This is achieved by aggregating alert rows based on shared IncidentIds from the alert dataframe containing all alerts, and summing their respective numerical columns. 
For incidents with multiple grades, the majority label is applied, with ties going to the true positive class.
We remove any incidents without a triage grade at this stage.

\begin{algorithm}[!t]
\KwIn{Alert data $\mathbf{A}$, minimum cardinality $c$, principal components $k$, max incidents sampled per IncidentHash $m$, max incidents stored per IncidentHash $s$, grid search parameters $\mathcal{G}$}
\KwOut{Incident embeddings $\mathbf{I}$ \& trained models $\mathcal{M}_{\text{t}}, \mathcal{M}_{\text{r}}$} 

\SetKwBlock{Feature}{Feature Engineering}{end}
\SetKwBlock{Training}{Model Training}{end}

\BlankLine
\Feature{
    $\mathbf{A} \leftarrow \text{FeatureEngineering}(\mathbf{A})$ \tcp*{T1} 
    $\mathbf{A} \leftarrow \text{FeatureSpaceCompression}(\mathbf{A}, c)$ \tcp*{T2} 
    $\mathbf{A} \leftarrow \text{OneHotEncoding}(\mathbf{A})$ \tcp*{T3} 
    $\mathbf{I} \leftarrow \text{FormIncidents}(\mathbf{A})$ \tcp*{T4} 
    $\mathbf{I} \leftarrow \text{SampleIncidents}(\mathbf{I}, m)$ \tcp*{T5} 
    $\mathbf{I}, \mathbf{A} \leftarrow \text{PCA}(\mathbf{I}, k), \text{PCA}(\mathbf{A}, k)$ \tcp*{T6} 

    $\text{StoreEmbeddings}(\mathbf{I}, s)$ \tcp*{T7}
}

\BlankLine

\Training{
    $\mathbf{I'}, \mathbf{A'} \leftarrow \text{ConvertToPandas}(\mathbf{I}, \mathbf{A})$ \tcp*{T8}

    $\mathcal{M}_t \leftarrow \text{TrainTriageModel}(\mathbf{I'}, \mathcal{G})$ \tcp*{T9}
    
    $\mathcal{M}_r \leftarrow \text{TrainRemediationModel}(\mathbf{A'}, \mathcal{G})$ \tcp*{T9}   

    $\text{ValidateAndStoreModels}(\mathcal{M}_t, \mathcal{M}_r)$ \tcp*{T10}
}

 \caption{\method{} Training}
 \label{alg:train}
\end{algorithm}

\medskip\noindent
\textbf{T5---Sampling incidents.}
Given that incident processing steps are significantly more memory-intensive than remediation actions in the alert dataframe, we employ random sampling on the incident dataframe to mitigate out-of-memory issues during downstream processing steps. 
This sampling strategy involves creating a unique IncidentHash identifier for each incident by arranging the DetectorIds of an incident into an ordered list and hashing it using SHA1. 
We can then cap the number of incidents for each unique IncidentHash and triage grade to a maximum of 1,000.

\medskip\noindent
\textbf{T6---Dimensionality reduction.}
We independently apply principal component analysis (PCA) to both the incident and alert dataframes, each containing tens of thousands of columns. 
PCA is chosen for its performane and availability within PySpark's native machine learning library, which supports distributed computing and allows for efficient feature space reduction. 
This mitigates the risk of out-of-memory errors that occur when centralizing large dataframes on the primary PySpark node for subsequent scikit-learn model training.
Our objective is to condense the feature space to 
$k$ principal components that captures 95\% of the original variance in each dataframe. 
Empirically, we find that setting $k=40$ meets this requirement. 
The resulting PCA weights are saved in an Azure Blob Storage container, enabling reuse within the inference and embedding pipelines.

\medskip\noindent
\textbf{T7---Store embeddings.}
The final step is to save the incident embeddings to an ADLS table to enhance similar incident recommendations within the inference pipeline. 
A strategic product decision dictates that only the top five most similar incidents are displayed for any given incident. 
Consequently, we only store up to five instances of each incident, categorized by triage grade and the unique set of DetectorIds that comprise the IncidentHash.

\subsection{Model Training}\label{subsec:train}
We train two models---a triage model to predict incident grades and an action model to determine remediation actions for incident-related alerts---across three key steps: 
(1) converting the PySpark dataframes to Pandas and performing a stratified train-val-test split; 
(2) optimizing random forest models with a grid search; 
and (3) validating new models against previous versions before storage.
While PySpark's native MLlib is an option, our experiments show it results in a 10\% decrease in Macro-F1 score compared to scikit-learn due to missing core capabilities. 
For instance, MLlib's random forest model is limited to a depth of 30, significantly constraining its ability to capture complex patterns.

\medskip\noindent
\textbf{T8---Dataset formation.}
We begin by converting the alert and incident PySpark dataframes created in Section~\ref{subsec:feature} into Pandas. 
For each Pandas dataframe, we conduct a standard 70-10-20 train, validation, and test set split of the data, stratified by grade and action labels, respectively.

\medskip\noindent
\textbf{T9---Training process.}
We select a random forest model due to its efficiency on our CPU-based PySpark infrastructure and its reliable performance on tabular data. 
We conduct a grid search over four key model parameters: $n\_estimators=\{100, 200, 300, 400\}$, $max\_depth=\{30, 50, 75, 100\}$, $min\_samples\_split=\{5, 10, 15\}$, and $class\_weight=\{'balanced', None\}$, and select the model with the highest macro-F1 score on the validation set.

\medskip\noindent
\textbf{T10---Validation and model storage.}
Once the best model has been identified for both triage and remediation tasks, we compare them with previous models to ensure quality between training cycles remains within a few percentage points. 
We do not require each model to have a higher macro-F1 score since new detectors and security products are onboarded over time, which can cause fluctuations in performance. 
After validation, the models are saved to an Azure Blob Storage container for use in the inference pipeline.

\section{Inference Pipeline}\label{sec:inference}
Building on the infrastructure in Section~\ref{sec:train}, the inference pipeline processes batched data to provide guided response recommendations (Sec~\ref{subsec:preprocessing}). 
For each new or updated incident, the trained models and historical embeddings operate across three phases: triage to predict grades (Sec~\ref{subsec:triage}), investigation to find similar incidents (Sec~\ref{subsec:investigation}), and remediation to recommend response actions (Sec~\ref{subsec:remediation}). 
Recommendations dynamically update as incidents evolve and are stored in a table, ensuring rapid access for customers. Algorithm~\ref{alg:train} provides a step-by-step overview.

\subsection{Preprocessing}\label{subsec:preprocessing}
The initial phase of the inference pipeline prepares real-time batched alert data, retrieving the last 15 minutes of telemetry and loading it into a PySpark dataframe. 
These alerts are then processed using the feature space compression and one-hot encoding techniques outlined in Section~\ref{subsec:feature}. 
Next, we bifurcate the alert data into two distinct PySpark dataframes---one dedicated to generating remediation predictions and the other for aggregating alerts into incidents for similar incident and triage recommendations, utilizing the aggregation process described in Section~\ref{subsec:feature}. 
Afterwards, we apply the latest PCA models from the training pipeline to reduce the dimensionality of both dataframes to form alert and incident embeddings. 
The incident embeddings are stored in an ADLS table to enhance the similar incident recommendations, with a limit of five instances per incident, categorized by triage grade and DetectorId.
Finally, we convert both PySpark dataframes into Pandas dataframes to facilitate subsequent triage, investigation, and remediation processes.

\begin{algorithm}[!t]
\KwIn{Batched alert data $\bm{A}$, max incidents stored per IncidentHash $s$, triage and remediation models $\mathcal{M}_{\text{t}}, \mathcal{M}_{\text{r}}$ with confidence thresholds $c_t$, $c_r$}
\KwOut{Triage, investigation, and remediation recs}

\BlankLine

\SetKwBlock{Preprocessing}{Preprocessing}{end}
\SetKwBlock{Triage}{Triage Recommendations}{end}
\SetKwBlock{Investigation}{Investigation Recommendations}{end}
\SetKwBlock{Remediation}{Remediation Recommendations}{end}

\Preprocessing{
    $\mathbf{A} \leftarrow \text{FeatureEngineering}(\mathbf{A})$ \tcp*{T1} 
    $\mathbf{A} \gets \text{FeatureSpaceCompression}(\mathbf{A})$ \tcp*{T2}
    
    $\mathbf{A} \gets \text{OneHotEncoding}(\mathbf{A})$ \tcp*{T3} 
    
    $\mathbf{I} \leftarrow \text{FormIncidents}(\mathbf{A})$ \tcp*{T4} 
    
    $\mathbf{I}, \mathbf{A} \leftarrow \text{PCA}(\mathbf{I}), \text{PCA}(\mathbf{A})$ \tcp*{T6} 
    
    $\text{StoreEmbeddings}(\mathbf{I}, s)$ \tcp*{T7}
    $\mathbf{I'}, \mathbf{A'} \gets \text{ConvertToPandas}(\mathbf{I}, \mathbf{A})$
}

\BlankLine

\Triage{
    $\mathcal{R}_\text{tri} \gets \mathcal{M}_\text{t}(\mathbf{I})$ \\
    $\mathcal{R}_\text{tri} \gets \text{FilterByConfidence}(\mathcal{R}_\text{tri}, c_t)$
}

\BlankLine

\Investigation{
    $\mathbf{H} \gets \text{RetrieveHistoricalEmbeddings}(180\text{ days})$ \\
    $\mathcal{R}_\text{inv} \gets \emptyset$ \\
    \ForEach{$\mathbf{I}_\text{new} \in \mathbf{I}_\text{tri}$}{
        $\mathcal{R}_\text{inv} \gets \mathcal{R}_\text{inv} \cup \text{ExactHashMatch}(\mathbf{I}_\text{new}, \mathbf{H})$ \\
        $\mathcal{R}_\text{inv} \gets \mathcal{R}_\text{inv} \cup \text{CosineSimilarityMatch}(\mathbf{I}_\text{new}, \mathbf{H})$ \\
        $\mathcal{R}_\text{inv} \gets \text{SelectTopK}(\mathcal{R}_\text{inv}, 5)$
    }
}

\BlankLine

\Remediation{
    $\mathcal{R}_\text{rem} \gets \mathcal{M}_\text{r}(\mathbf{A})$ \\
    $\mathcal{R}_\text{rem} \gets \text{FilterByConfidence}(\mathcal{R}_\text{rem}, c_r)$ \\
    $\mathcal{R}_\text{rem} \gets \text{IdentifyEntities}(\mathcal{R}_\text{rem})$ \\
    $\mathcal{R}_\text{rem} \gets \text{AggregateRecommendations}(\mathcal{R}_\text{rem}, \mathcal{I})$
}

\caption{\method{} Inference}
\label{alg:inference}
\end{algorithm}

\subsection{Triage Recommendations}\label{subsec:triage}
We leverage the incident embeddings produced during preprocessing, along with the latest version of the triage model, to generate triage recommendations. 
Each incident is evaluated by the model and given a prediction of true positive (TP), false positive (FP), or benign positive (BP), the latter considered as an informational incident.
The confidence of each recommendation is assessed against a precision threshold of $0.9$ to ensure that only reliable recommendations are sent to SOC analysts.

\subsection{Investigation Recommendations}\label{subsec:investigation}
We utilize the incident embeddings produced during the preprocessing step to generate recommendations for similar incidents. 
This process begins with the retrieval of historical incident embeddings from our Azure Data Lake Storage (ADLS), going back up to 180 days.
These embeddings capture past incidents in a vectorized format, enabling efficient comparison.
The core of our approach is to match new incidents with historically relevant incidents within the same organization through a three-step matching process:

\begin{enumerate}[topsep=4pt, leftmargin=*, itemsep=3pt]

    \item \textbf{Exact hash matching.} We begin by identifying historical incidents that share the same IncidentHash and triage recommendation. 
    If less than five matches are found, we take incidents with the same IncidentHash but differing triage recommendations. 

    \item \textbf{Approximate matching with cosine similarity.} If less than five exact matches were found, we search for historical incidents based on the cosine similarity of their embeddings. 
    This approach helps to identify incidents that share significant characteristics with the current incident.

    \item \textbf{Top-k similar incident selection.} We select the top-k most similar incidents, up to a maximum of five. 
    Exact and cosine similarity matches are ordered, with a higher priority given to exact matches to ensure the most germane comparisons, and ties going to the most recent incident.
\end{enumerate}

\subsection{Remediation Recommendations}\label{subsec:remediation}
Using alert embeddings from preprocessing and the latest remediation model, we generate targeted response actions---contain user, isolate machine, or stop virtual machine---for each alert with confidence above a $0.9$ precision threshold. 
The system identifies entities (e.g., users, devices, VMs) using encoded rules based on security domain knowledge, and then aggregates the individual alert recommendations into comprehensive incident recommendations.

\section{Embedding Pipeline}\label{sec:embed}
We generate historical incident embeddings that allow the similar incident recommendation algorithm to leverage up to 180 days of historical data when making recommendations. 
Due to the limitations of the training pipeline in processing huge volumes of incident telemetry across regions, we developed a specialized mechanism to generate historical embeddings. 
This ensures that our similar incident recommendation algorithm rapidly reaches comprehensive historical coverage each time the inference pipeline is executed.

\medskip\noindent
\textbf{Continuous embedding generation.}
The embedding pipeline operates in a continuous loop, with each iteration processing data one day further back than the last. 
Leveraging the preprocessing steps outlined in Section~\ref{subsec:preprocessing}, we integrate a deduplication process that loads historical incident embeddings and compares incident hashes and triage recommendations to eliminate redundancy.
We store any IncidentHash and triage recommendation pairs from the current batch, including those without a triage grade, provided they do not exceed five stored embeddings---aligning with our policy of recommending no more than five similar incidents at a time.
New incident embeddings are then saved to the ADLS table for use in the inference pipeline.
This procedure repeats until we have 180 days of historical incident embedding telemetry. 

\section{Experiments}\label{sec:experiments}
We evaluate \methodshort{}'s performance across three tasks: 
(1) triaging, assessing the model's ability to classify incidents as benign, malicious, or informational; 
(2) investigation, analyzing the relevance of similar incident recommendations; 
and (3) remediation, evaluating its ability to predict effective threat mitigation actions.

\begin{table*}[t]
\centering
\begin{tabular}{lrrrrrrrrrrrrrrrrrrr} 
\multicolumn{4}{c}{\textbf{Train Statistics}} & \multicolumn{7}{c}{\textbf{Triage Results}} & \multicolumn{7}{c}{\textbf{Remediation Results}} \\
\cmidrule(lr){1-4} \cmidrule(lr){5-11} \cmidrule(lr){12-18}
 \textbf{Region} &  \textbf{\# Rules} &  \textbf{\# Alerts} &  \textbf{\# Inc} & \textbf{Supp} & \textbf{\% TP} & \textbf{\% FP} & \textbf{\% BP} &\textbf{Pr} & \textbf{Re} & \textbf{F1} & \textbf{Supp} & \textbf{\% CA} & \textbf{\% ID} & \textbf{\% VM} & \textbf{Pr} & \textbf{Re} & \textbf{F1}   \\
\midrule
      1 & 31k & 18.4M & 5.1M & 49k & 16 & 32 & 52 & .86 & .85 & .85 & 113k & 86 & 14 & - & 1 & 1 & 1  \\
      
      2 & 31k & 14.5M & 4.7M & 46k & 14 & 34 & 52 & .92 & .90 & .91 & 166k & 86 & 14 & 1 & 1 & 1 & 1  \\
      
      3 & 26k & 11.9M & 4.3M & 97k & 21 & 25 & 54 & .85 & .82 & .84 & 180k & 86 & 14 & - & .93 & .99 & .96  \\
      
      4 & 18k & 9.2M & 3.2M & 83k & 23 & 29 & 48 &.87 & .85 & .86 & 82k & 88 & 12 & - & 1 & 1 & 1  \\
      
      5 & 14k& 6M & 1.9M & 99k & 20 & 38 & 43 & .88 & .86 & .86 & 46k & 90 & 10 & - & 1 & 1 & 1  \\
      
      6 & 15k & 6M & 2M & 116k & 23 & 38 & 39 & .87 & .87 & .87 & 21k & 73 & 24 & 2 & 1 & 1 & 1   \\
      
      7 & 6.7k & 2.5M & 748k & 138k & 29 & 39 & 32 & .86 & .86 & .86 & 11k & 69 & 31 & - & 1 & 1 & 1  \\
      
      8 & 6.8k & 2.1M & 657k & 139k & 21 & 33 & 46 & .88 & .87 & .88 & 9.8k & 63 & 37 & - & 1 & 1 & 1  \\
      
      9 & 6.3k & 1.7M & 744k & 87k & 13 & 17 & 70 & .90 & .85 & .87 & 12k & 84 & 16 & - & .99 & .99 & 1   \\
      
     10 & 3k & 511k & 250k & 106k & 14 & 38 & 48 & .88 & .86 & .87 & 1.9k & 45 & 55 & - & 1 & 1 & 1  \\
     
     11 & 798 & 109k & 71k & 23k & 20 & 52 & 28 & .84 & .85 & .84 & - & - & - & - & - & - & - \\
     
     12 & 579 & 25k & 10k & 8.9k & 17 & 46 & 37 & .87 & .88 & .87 & 270 & 45 & 55 & - & 1 & 1 & 1 \\
\bottomrule
\end{tabular}
\caption{
Left: Statistics on the number of unique DetectorIds, volume of alerts, and volume of incidents across 12 sampled regions over two-weeks. 
Middle: Triage model statistics, including the number of graded incidents (``Supp''), the distribution of grades (i.e., TP, FP, and BP), and model performance metrics evaluated through macro precision, recall, and F1 score. 
The triage statistics are sourced from numerous first and third party providers, with a large proportion of BP and FP incidents coming from third party providers.
Right: Remediation model statistics, including the number of alerts with an action label (``Supp''), the distribution of contain account, isolate device, and stop virtual machine actions, and model performance metrics.
}
\label{table:combined-stats}
\vspace{-5mm}
\end{table*}

\subsection{Setup}
We present results for the triage and remediation models across a sample of 12 regions, where each regional dataset is divided into three stratified subsets: training (70\%), validation (10\%), and testing (20\%). 
Each run of model training job includes two key parameter optimizations, performed independently for incident triage and alert remediation embeddings:

\begin{enumerate}[topsep=2mm, itemsep=0mm, parsep=1mm, leftmargin=*]
    \item \textbf{PCA component selection.} We retain the top-k principal components capturing 95\% of the data variance, with $k$ ranging up to $100$. 
    While the optimal $k$ varies across runs and regions, we find that $k=40$ performs well across most scenarios.

    \item \textbf{Random forest parameter tuning.} We optimize key random forest parameters using a grid search (see Section~\ref{subsec:train}) 
    We select the model with the highest macro-F1 score on the validation set and report precision and recall metrics, as is standard for imbalanced datasets~\cite{freitas2022malnet,freitas2021large,duggal2021har,duggal2020rest}.
\end{enumerate}

\begin{figure}[b]
    \centering
    \includegraphics[width=\linewidth]{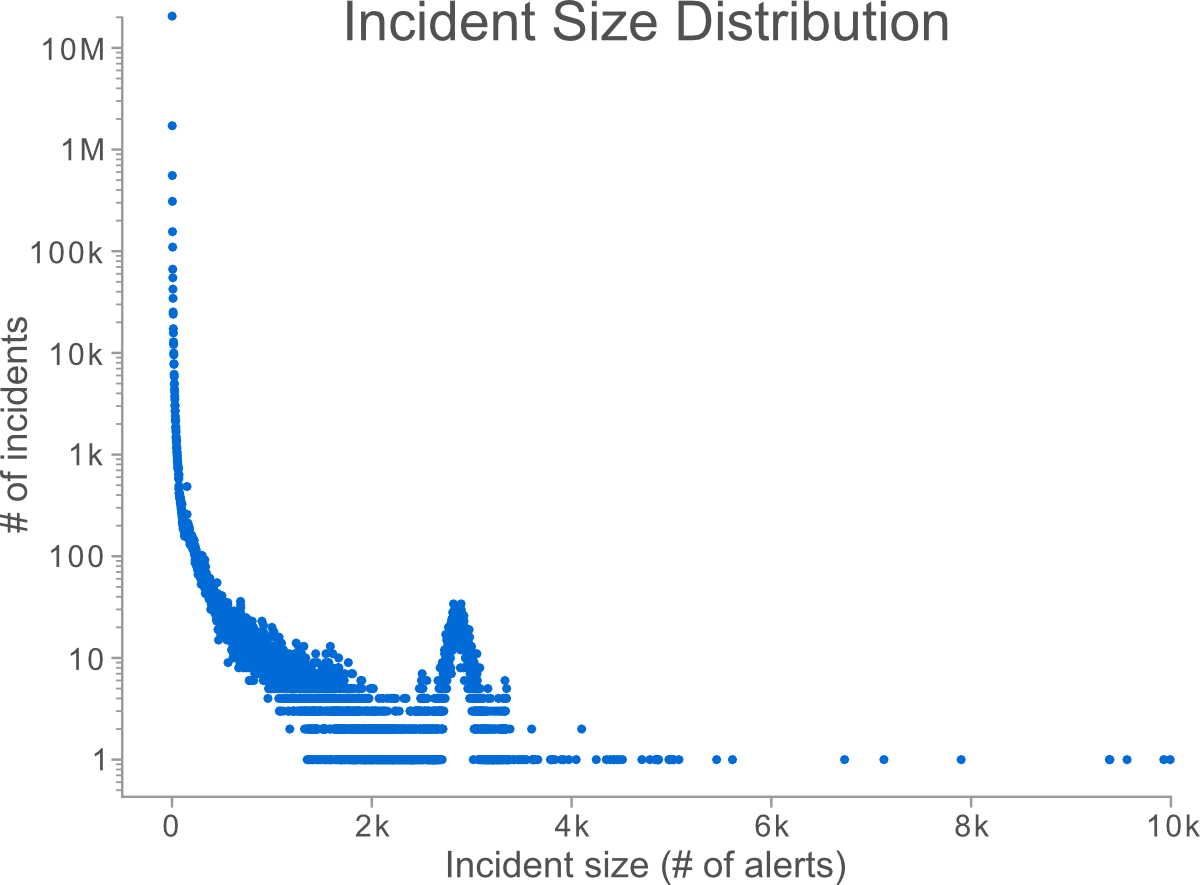}
    \caption{Sampled distribution of incident size---measured by the number of alerts per incident---exhibits a long-tailed pattern where the majority of incidents have only a few alerts.}
    \label{fig:incident-distribution}
\end{figure}

Table~\ref{table:combined-stats} summarizes model training and performance statistics over two weeks across select regions, including the count of unique DetectorIds, and alert/incident volumes.
Regional variations are significant, with up to 31k detectors across thousands of organizations, complicating the training process. 
Incident size distribution (Figure~\ref{fig:incident-distribution}) is long-tailed, with most incidents comprised of a few alerts, with larger incidents representing a greater challenge for triage and similar incident recommendations due to their rarity.

\smallskip\noindent
\textbf{Limitations.}
This work focuses on presenting a scalable, unified framework for incident triaging, remediation, and similar incident recommendations, rather than evaluating alternative models or competing security products, which are often undisclosed.
The release of the \dataset{} dataset enables researchers to explore and optimize new architectures for maximum performance.
In addition, \methodshort{} is limited to providing recommendations for existing detectors and does not address zero-day or other unmonitored attack vectors.

\subsection{\dataset{} Dataset}
\dataset{} includes over 13M pieces of evidence across 33 entity types, encompassing 1.6M alerts and 1M incidents over a two-week period. 
Each incident is annotated with a ground-truth triage label by customer security analysts, along with 26k alerts containing labels of remediation actions taken by customers.
The dataset is derived from Region 2 
(see Table~\ref{table:combined-stats}) and includes telemetry across 6.1k organizations, featuring 9.1k unique custom and built-in DetectorIds across numerous security products, covering 441 MITRE ATT\&CK techniques~\cite{strom2018mitre}. 
We divide the dataset into a train (70\%) and test set (30\%), both available as CSV files on \href{https://www.kaggle.com/datasets/Microsoft/microsoft-security-incident-prediction}{Kaggle}. 
The release of \dataset{} provides an unparalleled opportunity for the development of GR systems and beyond, with \methodshort{} providing a foundational baseline. 
Additional details on \dataset{} can be found in the Appendix.

\subsection{Triage}
The telemetry collection period for triage ranges from 7 to 180 days, varying by region due to computational constraints in incident preprocessing.
Table~\ref{table:combined-stats} shows that the number of SOC-graded incidents (abbreviated as "Supp" for support) ranges from 8.9k to 139k with an imbalance among the three triage classes of 19\% true positive, 35\% false positive, and 46\% benign positive (informational).
While larger regions tend to have more graded incidents, several factors complicate the number of graded incidents reported during training: 
(1) larger regions with more organizations and detectors require more memory during OHE, reducing capacity for additional alerts; 
(2) the frequency and types of grading vary significantly across SOCs and regions,
and (3) we limit the number of graded examples per IncidentHash and triage label to 1k.

\medskip\noindent
\textbf{Offline results.}
In Table~\ref{table:combined-stats}, we present the cross-region model triage performance at the point of maximum macro-F1 score on the precision-recall curve.
The results show an average macro-F1 score of $0.87$, with a precision of $0.87$, and a recall of $0.86$. 
This performance demonstrates the model's ability to effectively manage the complexities of the triage task, ranging from incidents involving a single alert to those comprising hundreds of alerts across numerous detectors and security products.

\begin{figure}[b]
    \centering
    \includegraphics[width=\linewidth]{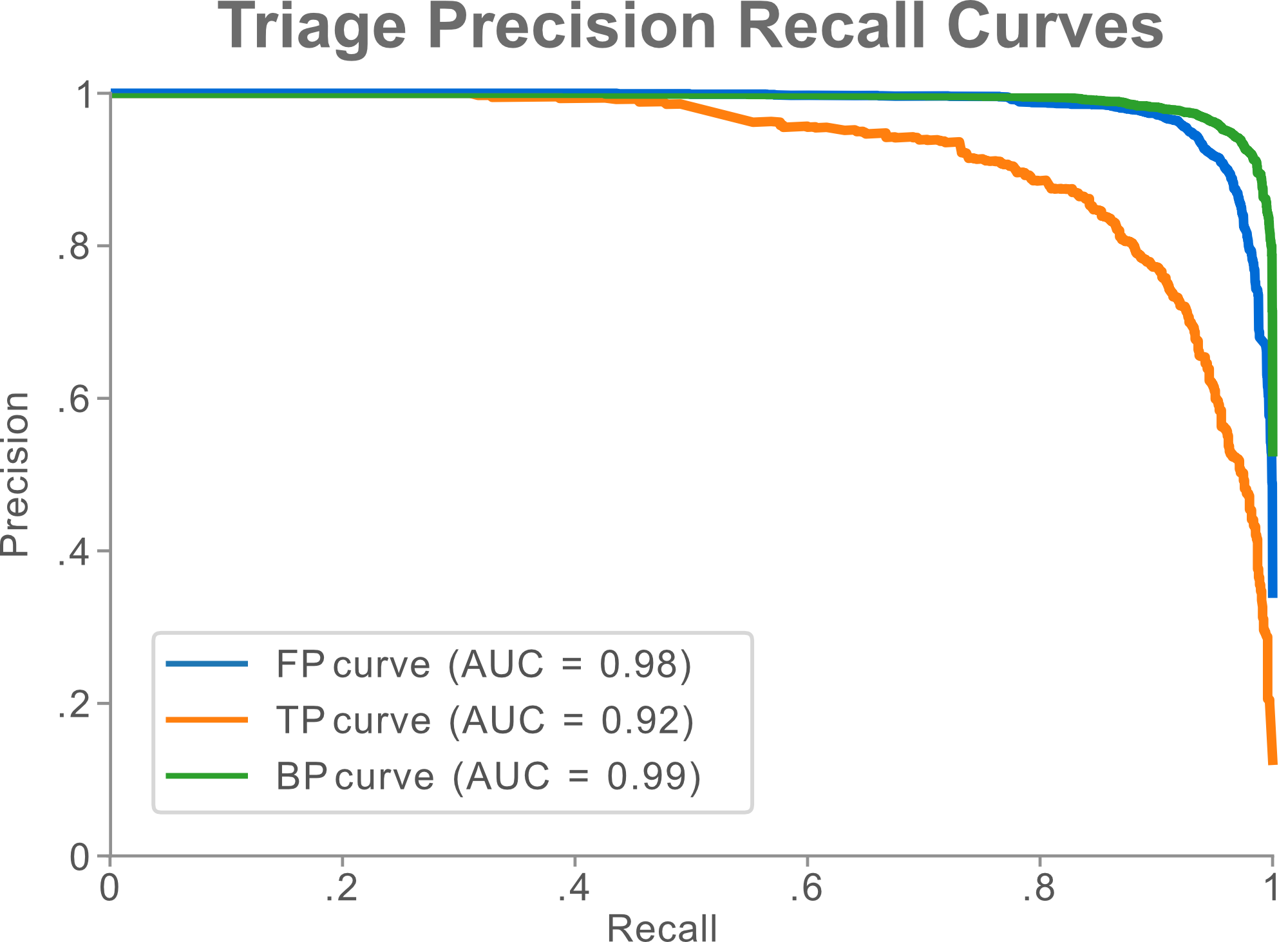}
    \caption{PR curves of triage performance in Region 2}
    \label{fig:pr-curves}
\end{figure}

To evaluate \methodshort's effectiveness at a more granular level, we examine performance across individual triage classifications in Region 2. 
The precision-recall curves in Figure~\ref{fig:pr-curves} show that \methodshort{} consistently achieves high AUC scores across triage labels.
The majority of misclassifications (81.1\%) occur when incidents are categorized as BP instead of TP or FP, or vice versa—a discrepancy mainly due to the lack of a standardized definition for BP incidents, leading to classification inconsistencies across and within SOCs. 
However, the more critical incident misclassification of TP as BP or FP is rare (2.4\%), as shown in Table~\ref{tab:confusion-matrix}.

\medskip\noindent
\textbf{Online results.}
Triage recommendations are inherently dynamic, adapting as new alerts are added to incidents.
However, as shown in Figure~\ref{fig:incident-distribution}, since the vast majority of incidents are relatively short-lived and involve only a few alerts, only 2\% of incidents undergo a change in their initial triage recommendation.
Furthermore, model training pipelines across regions exhibit some bias, as not all types of incidents and alerts are graded by SOC analysts. 
Coupled with a heightened precision threshold to ensure that recommendations are correct $90\%$ of the time, the triage models effectively cover 41\% of incidents across regions.

\begin{table}[t]
\setlength{\tabcolsep}{12pt}
    \centering
    \begin{tabular}{c|ccc}
         \textbf{Actual / Predicted} & \textbf{TP} & \textbf{BP} & \textbf{FP} \\
        \hline
        \textbf{TP} & 929 & 170 & 50 \\
        \textbf{BP} & 62 & 4,748 & 45 \\
        \textbf{FP} & 66 & 221 & 2,869 \\
    \end{tabular}
    \caption{Confusion matrix of triage performance in Region 2.
    Diagonal values reflect correct classifications, highlighting strong model performance across triage classes.}
    \label{tab:confusion-matrix}
\vspace{-3mm}
\end{table}

\subsection{Investigation}
In collaboration with security experts at Microsoft, we manually evaluated our similar incident recommendations due to an absence of a definitive ground truth. 
For this evaluation, we randomly selected 1k incidents varying in size, detectors, products, organizations, and regions. 
Security researchers were then tasked with randomly selecting a similar incident recommendation for each reference incident, and judging its relevance based on criteria such as shared attack patterns, indicators of compromise, and entity types. 

\medskip\noindent
\textbf{Offline results.}
The assessment found that 94\% of the recommended incidents were relevant, with only 2\% deemed dissimilar.
The quality of recommendations tends to diminish for smaller organizations with limited historical data.
Similarly, larger incidents characterized by hundreds of alerts across multiple products and detectors also show decreased recommendation quality due to their rarity.
To address this, a cosine similarity threshold of $0.9$ was identified as the cut-off threshold to ensure that only relevant recommendations are presented to customers.
Across regions, we find that 98\% of all incidents have one or more recommendations.

\subsection{Remediation}
We collect 180 days of telemetry, utilizing a preprocessing pipeline for alerts that is significantly simpler than the one used for incident-based triage. 
Unlike incidents, alerts do not require sampling due to their less computationally intensive preprocessing, allowing for a higher volume of training data, particularly in larger regions as shown in Table~\ref{table:combined-stats}. 
The volume of actioned alerts varies widely across regions, from a few hundred to 180k, with a notable imbalance among the remediation classes---67\% are contain account (CA), 23\% isolate device (ID), and less than 1\% stopping virtual machines (VM). 
Future work to support remediation actions, such as quarantining files, deleting emails, and blocking IPs/URLs is ongoing.

\medskip\noindent
\textbf{Offline results.}
We achieve an impressive average cross-region macro-F1 score of $0.99$. 
This high score reflects the relative simplicity of predicting the appropriate remediation action for a single alert compared to the complexities of incident triaging in our dataset.

\medskip\noindent
\textbf{Online results.}
While we achieve notable offline results, the  collected data does not fully capture the SOC experience. 
A majority of alerts are not actioned by analysts, and not all alerts fit the three predefined remediation action types. 
As a result, the model's coverage averages 62\% across regions.
However, as we integrate new remediation actions, the system's coverage will naturally increase.

\section{Deployment}\label{sec:deployment}
\method{} has been successfully deployed across the world, serving thousands of Microsoft Defender XDR customers since its launch in April 2024. 
\methodshort{} has generated millions of guided response recommendations for triage, investigation, and remediation tasks, receiving a `positive' user response rate of 89\%, based on the confirmation or dismissal of recommendations.

Our deployment infrastructure leverages a Synapse-based PySpark cluster, customized to each geographical region. 
This infrastructure includes:
(a) an ADLS database ensuring both accessibility and secure management of telemetry;
(b) an Azure Synapse backend that provides a robust framework for deployment;
(c) an XXL PySpark pool featuring 60 executors, each equipped with 64 CPU cores and 400GB of RAM;
(d) autoscaling to adjust executors based on fluctuating load;
and (e) automated re-execution of failed jobs to ensure continuous coverage.
Due to the absence of native support for model monitoring, versioning, and storage within Synapse, we developed a custom infrastructure to support these capabilities.

\section{Conclusion}\label{sec:conclusion}
\method{} (\methodshort{}) 
represents the first time a cybersecurity company has openly discussed an industry-scale guided response framework. 
\methodshort{} significantly enhances SOC operations by guiding security analysts through crucial investigation, triaging, and remediation tasks, adeptly handling everything from simple alerts to complex incidents.
The performance of \methodshort{} has been rigorously evaluated through internal testing, collaboration with Microsoft security experts, and extensive customer feedback, demonstrating its effectiveness across all three tasks. 
Deployed globally within Microsoft Defender XDR, \methodshort{} generates millions of guided response recommendations weekly, with 89\% of user interactions receiving positive feedback.
In addition, we release \href{https://www.kaggle.com/datasets/Microsoft/microsoft-security-incident-prediction}{\dataset{}}, the largest publicly available collection of real-world security incidents, comprising 13 million pieces of evidence across one million incidents, each annotated with ground-truth triage labels by customer security analysts.
As the first resource of its kind, \dataset{} sets a new standard for advancing the development and evaluation of guided response systems and beyond.
\begin{acks}
We thank our colleagues who supported this research, including Shira Shacham, Yuval Derman, Oren Saban, Noa Bratman, Inbar Rotem, Omri Kantor, Itamar Karavani, Mari Mishel, Yuval Katav, Niv Zohar, Lior Camri, Leeron Luzzatto, Anna Karp, Ido Nitzan, Corina Feurstein, Ya’ara Cohen, Nadia Tkach Mendes, Gil Shmaya, Pawel Partyka, Shachaf Levy, Blake Strom, along with many others.
\end{acks}

\bibliographystyle{ACM-Reference-Format}
\bibliography{main.bib}

\clearpage
\renewcommand{\thesubsection}{\Alph{subsection}}
\section*{Appendix}\label{sec:appendix}
\appendix

\subsection{Dataset Overview}
\dataset{} represents the largest publicly available collection of real-world cybersecurity incidents, containing over 13 million pieces of evidence across 1.6 million alerts and 1M annotated incidents.
Under the permissive CDLA-2.0 license, this dataset offers a unique opportunity for researchers and practitioners to develop and benchmark advanced ML models on authentic and comprehensive security telemetry. 
See Table~\ref{table:feature-descriptions} for a description of each dataset field.

We provide three hierarchies of data: (1) evidence, (2) alert, and (3) incident. 
At the bottom level, evidence supports an alert. 
For example, an alert may be associated with multiple pieces of evidence such as an IP address, email, and user details, each containing specific supporting metadata. 
Above that, we have alerts that consolidate multiple pieces of evidence to signify a potential security incident. 
These alerts provide a broader context by aggregating related evidences to present a more comprehensive picture of the potential threat. 
At the highest level, incidents encompass one or more alerts, representing a cohesive narrative of a security breach or threat scenario.

\subsection{Benchmarking}
With the release of \dataset{}, we aim to establish a standardized benchmark for guided response systems using real-world data. 
The primary objective of the dataset is to accurately predict incident triage grades—true positive (TP), benign positive (BP), and false positive (FP)—based on historical customer responses. 
To support this, we provide a training dataset containing 45 features, labels, and unique identifiers across 1M triage-annotated incidents. 
We divide the dataset into a train set containing 70\% of the data and a test set with 30\%, stratified based on triage grade ground-truth, OrgId, and DetectorId. 
We ensure that incidents are stratified together within the train and test sets to ensure the relevance of evidence and alert rows.
The CSV files are hosted on Kaggle: \url{https://www.kaggle.com/datasets/Microsoft/microsoft-security-incident-prediction}

A secondary objective of \dataset{} is to benchmark the remediation capabilities of guided response systems.
To this end, we release 26k ground-truth labels for predicting remediation actions for alerts, available at both granular and aggregate levels. 
The recommended metric for evaluating research using the \dataset{} dataset is macro-F1 score, along with details on precision and recall.

\subsection{Privacy}
To ensure privacy, we implement a stringent anonymization process. 
Initially, sensitive values are pseudo-anonymized using SHA1 hashing techniques. 
This step ensures that unique identifiers are obfuscated while maintaining their uniqueness for consistency across the dataset. 
Following this, we replace these hashed values with randomly generated IDs to further enhance anonymity and prevent any potential re-identification. 
Additionally, we introduce noise to the timestamps, ensuring that the temporal aspects of the data cannot be traced back to specific events. 
This multi-layered approach, combining pseudo-anonymization and randomization, safeguards the privacy of all entities involved while maintaining the integrity and utility of the dataset for research and development purposes.

\begin{table}[H]
\renewcommand*{\arraystretch}{0.87}
\centering
\small
\begin{tabular}{p{0.15\textwidth} p{0.295\textwidth}}

\textbf{Feature} & \textbf{Description} \\ 
\cmidrule(r){1-1} \cmidrule(lr){2-2}

Id & Unique ID for each OrgId-IncidentId pair \\ \addlinespace
OrgId & Organization identifier \\ \addlinespace
IncidentId & Organizationally unique incident identifier \\ \addlinespace
AlertId & Unique identifier for an alert \\ \addlinespace
Timestamp & Time the alert was created \\ \addlinespace
DetectorId & Unique ID for the alert generating detector \\ \addlinespace
AlertTitle & Title of the alert \\ \addlinespace
Category & Category of the alert \\ \addlinespace
MitreTechniques & MITRE ATT\&CK techniques involved in alert \\ \addlinespace
IncidentGrade & SOC grade assigned to the incident \\ \addlinespace
ActionGrouped & SOC alert remediation action (high level) \\ \addlinespace
ActionGranular & SOC alert remediation action (fine-grain) \\ \addlinespace
EntityType & Type of entity involved in the alert \\ \addlinespace
EvidenceRole & Role of the evidence in the investigation \\ \addlinespace
Roles & Additional metadata on evidence role in alert \\ \addlinespace
DeviceId & Unique identifier for the device \\ \addlinespace
DeviceName & Name of the device \\ \addlinespace
Sha256 & SHA-256 hash of the file \\ \addlinespace
IpAddress & IP address involved \\ \addlinespace
Url & URL involved \\ \addlinespace
AccountSid & On-premises account identifier \\ \addlinespace
AccountUpn & Email account identifier \\ \addlinespace
AccountObjectId & Entra ID account identifier \\ \addlinespace
AccountName & Name of the on-premises account \\ \addlinespace
NetworkMessageId & Org-level identifier for email message \\ \addlinespace
EmailClusterId & Unique identifier for the email cluster \\ \addlinespace
RegistryKey & Registry key involved \\ \addlinespace
RegistryValueName & Name of the registry value \\ \addlinespace
RegistryValueData & Data of the registry value \\ \addlinespace
ApplicationId & Unique identifier for the application \\ \addlinespace
ApplicationName & Name of the application \\ \addlinespace
OAuthApplicationId & OAuth application identifier \\ \addlinespace
ThreatFamily & Malware family associated with a file \\ \addlinespace
FileName & Name of the file \\ \addlinespace
FolderPath & Path of the file folder \\ \addlinespace
ResourceIdName & Name of the Azure resource \\ \addlinespace
ResourceType & Type of Azure resource \\ \addlinespace
OSFamily & Family of the operating system \\ \addlinespace
OSVersion & Version of the operating system \\ \addlinespace
AntispamDirection & Direction of the antispam filter \\ \addlinespace
SuspicionLevel & Level of suspicion \\ \addlinespace
LastVerdict & Final verdict of threat analysis \\ \addlinespace
CountryCode & Country code evidence appears in \\ \addlinespace
State & State of evidence appears in \\ \addlinespace
City & City evidence appears in \\ \addlinespace
\bottomrule
\end{tabular}
\caption{Description of each column in the \dataset{} dataset}
\label{table:feature-descriptions}
\end{table}

\end{document}